# An Efficient Algorithm for Surface Generation

Christer Samuelsson[*]
Universität des Saarlandes, FR 8.7, Computerlinguistik
Posfach 1150, D-66041 Saarbrücken, Germany
Internet: christer@coli.uni-sb.de

## Abstract

A method is given that "inverts" a logic grammar and displays it from the point of view of the logical form, rather than from that of the word string. LR-compiling techniques are used to allow a recursive-descent generation algorithm to perform "functor merging" much in the same way as an LR parser performs prefix merging. This is an improvement on the semantic-head-driven generator that results in a much smaller search space. The amount of semantic look-ahead can be varied, and appropriate tradeoff points between table size and resulting nondeterminism can be found automatically.

## 1 Introduction

With the emergence of fast algorithms and optimization techniques for syntactic analysis, such as the use of explanation-based learning in conjunction with LR parsing, see [Samuelsson & Rayner 1991] and subsequent work, surface generation has become a major bottleneck in NLP systems. Surface generation is the inverse problem of syntactic analysis and subsequent semantic interpretation. The latter consists in constructing some semantic representation of an input word-string based on the syntactic and semantic rules of a formal grammar. In this article, we will limit ourselves to logic grammars that attribute word strings with expressions in some logical formalism represented as terms with a functor-argument structure. The surface generation problem then consists in assigning an output word-string to such a term. In general, both these mappings are many-to-many: A word string that can be mapped to several distinct logical forms is said to be ambiguous. A logical form that can be assigned to several different word strings is said to have multiple paraphrases.

We want to create a generation algorithm that generates a word string by recursively descending through a logical form, while delaying the choice of grammar rules to apply as long as possible. This means that we want to process different rules or rule combinations that introduce the same piece of semantics in parallel until they branch apart. This will reduce the amount of spurious search, since we will gain more information about the rest of the logical form before having to commit to a particular grammar rule.

In practice, this means that we want to perform "functor merging" much in the same ways as an LR parser performs prefix merging by employing parsing tables compiled from the grammar. One obvious way of doing this is to use LR-compilation techniques to compile generation tables. This will however require that we reformulate the grammar from the point of view of the logical form, rather than from that of the word string from which it is normally displayed.

This gives us the following working plan: We will first review basic LR compilation of parsing tables in Section 2. The grammar-inversion procedure turns out to be most easily explained in terms of the semantic-head-driven generation (SHDG) algorithm. We will therefore proceed to outline the SHDG algorithm in Section 3. The grammar inversion itself is described in Section 4, while LR compilation of generation tables is discussed in Section 5. The generation algorithm is presented in Section 6 together with techniques for optimizing the generation tables. Section 7, finally, discusses the findings.

## 2 LR Compilation for Parsing

LR compilation in general is well-described in for example [Aho et al. 1986], pp. 215–247. Here we will only sketch out the main ideas.

An LR parser is basically a pushdown automaton, i.e., it has a pushdown stack in addition to a finite set of internal states and a reader head for scanning the input string from left to right one symbol at a time. The stack is used in a characteristic way: The items on the stack consist of alternating grammar symbols and states. The current state is simply the state on top of the stack. The most distinguishing feature of an LR parser is however the form of the transition relation — the action and goto tables. A nondeterministic LR parser can in each step perform one of four basic actions. In state $S$ with lookahead symbol[1] *Sym* it can:

---

[*]The work presented in this article was funded by the N3 "Bidirektionale Linguistische Deduktion (BiLD)" project in the Sonderforschungsbereich 314 *Künstliche Intelligenz — Wissensbasierte Systeme*.

[1]The lookahead symbol is the next symbol in the input string, i.e., the symbol under the reader head.

| | | | |
|---|---|---|---|
| $S$ | $\rightarrow$ | $S\ QM$ | 1 |
| $S$ | $\rightarrow$ | $NP\ VP$ | 2 |
| $VP$ | $\rightarrow$ | $VP\ PP$ | 3 |
| $VP$ | $\rightarrow$ | $VP\ AdvP$ | 4 |
| $VP$ | $\rightarrow$ | $V_i$ | 5 |
| $VP$ | $\rightarrow$ | $V_t\ NP$ | 6 |
| $PP$ | $\rightarrow$ | $P\ NP$ | 7 |

| | | |
|---|---|---|
| $NP$ | $\rightarrow$ | *John* |
| $NP$ | $\rightarrow$ | *Mary* |
| $NP$ | $\rightarrow$ | *Paris* |
| $V_i$ | $\rightarrow$ | *sleeps* |
| $V_t$ | $\rightarrow$ | *sees* |
| $P$ | $\rightarrow$ | *in* |
| $AdvP$ | $\rightarrow$ | *today* |
| $QM$ | $\rightarrow$ | *?* |

Figure 1: Sample grammar

1. **accept**: Halt and signal success.
2. **error**: Fail and backtrack.
3. **shift** $S_2$: Consume the input symbol *Sym*, push it onto the stack, and transit to state $S_2$ by pushing it onto the stack.
4. **reduce** $R$: Pop off two items from the stack for each phrase in the RHS of grammar rule $R$, inspect the stack for the old state $S_1$ now on top of the stack, push the *LHS* of rule $R$ onto the stack, and transit to state $S_2$ determined by *goto($S_1$,LHS,$S_2$)* by pushing $S_2$ onto the stack.

Consider the small sample grammar given in Figure 1. To make this simple grammar more interesting, the addition of a question mark ($QM$) to the end of a sentence ($S$), as in *John sleeps?*, is interpreted as a yes-no question version of $S$ by the recursive Rule 1, $S \rightarrow S\ QM$.

Each internal state consists of a set of dotted items. Each item in turn corresponds to a grammar rule. The current string position is indicated by a dot. For example, Rule 2, $S \rightarrow NP\ VP$, yields the item $S \Rightarrow NP \cdot VP$, which corresponds to just having found an *NP* and now searching for a *VP*.

In the compilation phase, new states are induced from old ones: For the indicated string position, a possible grammar symbol is selected and the dot is advanced one step in all items where this particular grammar symbol immediately follows the dot, and the resulting new items will constitute the kernel of the new state. Non-kernel items are added to these by selecting grammar rules whose LHS match grammar symbols at the new string position in the new items. In each non-kernel item, the dot is at the beginning of the rule. If a set of items is constructed that already exists, then this search branch is abandoned and the recursion terminates.

The state-construction phase starts off by creating an initial set consisting of a single dummy kernel item and its non-kernel closure. This is State 1 in Figure 2. The dummy item introduces a dummy top grammar symbol as its LHS, while the RHS consists of the old top symbol, and the dot is at the beginning of the rule. In the example, this is the item $S' \Rightarrow \cdot S$. The rest of the states are induced from the initial state. The states resulting from the sample grammar of Figure 1 are shown in Figure 2.

In conjunction with grammar formalisms employing complex feature structures, this procedure is associated with a number of interesting problems, many of which are discussed in [Nakazawa 1991] and [Samuelsson 1994]. For example, the termination criterion must be modified:

State 1
$S' \Rightarrow \cdot S$
$S \Rightarrow \cdot S\ QM$
$S \Rightarrow \cdot NP\ VP$

State 2
$S' \Rightarrow S \cdot$
$S \Rightarrow S \cdot QM$

State 3
$S \Rightarrow NP \cdot VP$
$VP \Rightarrow \cdot VP\ PP$
$VP \Rightarrow \cdot VP\ AdvP$
$VP \Rightarrow \cdot V_i$
$VP \Rightarrow \cdot V_t\ NP$

State 4
$S \Rightarrow NP\ VP \cdot$
$VP \Rightarrow VP \cdot PP$
$VP \Rightarrow VP \cdot AdvP$
$PP \Rightarrow \cdot P\ NP$

State 5
$VP \Rightarrow VP\ PP \cdot$

State 6
$VP \Rightarrow VP\ AdvP \cdot$

State 7
$PP \Rightarrow P \cdot NP$

State 8
$PP \Rightarrow P\ NP \cdot$

State 9
$VP \Rightarrow V_i \cdot$

State 10
$VP \Rightarrow V_t \cdot NP$

State 11
$VP \Rightarrow V_t\ NP \cdot$

State 12
$S \Rightarrow S\ QM \cdot$

Figure 2: LR-parsing states for the sample grammar

If a new set of items is constructed that is *more specific* than an existing one, then this search branch is abandoned and the recursion terminates. If it on the other hand is *more general*, then it replaces the old one.

## 3 Semantic-Head-Driven Generation

Generators found in large-scale systems such as the DFKI DISCO system, [Uszkoreit *et al.* 1994], or the SRI Core Language Engine, [Alshawi (ed.) 1992], pp. 268–275, tend typically to be based on the semantic-head-driven generation (SHDG) algorithm. The SHDG algorithm is well-described in [Shieber *et al.* 1990]; here we will only outline the main features.

The grammar rules of Figure 1 have been attributed with logical forms as shown in Figure 3. The notation has been changed so that each constituent consists of a quadruple $\langle Cat, \text{Sem}, W_0, W_1 \rangle$, where $W_0$ and $W_1$ form a difference list representing the word string that *Cat* spans, and *Sem* is the logical form. For example, the logical form corresponding to the LHS $S$ of the $\langle S, \text{mod}(\text{X},\text{Y}), W_0, W \rangle \rightarrow \langle S, \text{X}, W_0, W_1 \rangle\ \langle QM, \text{Y}, W_1, W \rangle$ rule, consists of a modifier Y added to the logical form X of the RHS $S$. As we can see from the last grammar rule, this modifier is in turn realized as ynq.

For the SHDG algorithm, the grammar is divided into chain rules and non-chain rules: Chain rules have a distinguished RHS constituent, the semantic head, that has the same logical form as the LHS constituent, modulo $\lambda$-abstractions; non-chain rules lack such a constituent. In particular, lexicon entries are non-chain rules, since they do not have any RHS constituents at all. This distinction is made since the generation algorithm treats the two rule types quite differently. In the example grammar, rules 2 and 5 through 7 are chain rules, while the remaining ones are non-chain rules.

$\langle S, \mathtt{mod}(\mathtt{X},\mathtt{Y}), W_0, W \rangle \rightarrow$  1
  $\langle S, \mathtt{X}, W_0, W_1 \rangle \ \langle QM, \mathtt{Y}, W_1, W \rangle$
$\langle S, \mathtt{Y}, W_0, W \rangle \rightarrow$  2
  $\langle NP, \mathtt{X}, W_0, W_1 \rangle \ \langle VP, \mathtt{X\hat{}Y}, W_1, W \rangle$
$\langle VP, \mathtt{X\hat{}mod}(\mathtt{Y},\mathtt{Z}), W_0, W \rangle \rightarrow$  3
  $\langle VP, \mathtt{X\hat{}Y}, W_0, W_1 \rangle \ \langle AdvP, \mathtt{Z}, W_1, W \rangle$
$\langle VP, \mathtt{X\hat{}mod}(\mathtt{Y},\mathtt{Z}), W_0, W \rangle \rightarrow$  4
  $\langle VP, \mathtt{X\hat{}Y}, W_0, W_1 \rangle \ \langle PP, \mathtt{Z}, W_1, W \rangle$
$\langle VP, \mathtt{X}, W_0, W \rangle \rightarrow$  5
  $\langle V_i, \mathtt{X}, W_0, W \rangle$
$\langle VP, \mathtt{Y}, W_0, W \rangle \rightarrow$  6
  $\langle V_t, \mathtt{X\hat{}Y}, W_0, W_1 \rangle \ \langle NP, \mathtt{X} W_1, W \rangle$
$\langle PP, \mathtt{Y}, W_0, W \rangle \rightarrow$  7
  $\langle P, \mathtt{X\hat{}Y}, W_0, W_1 \rangle \ \langle NP, \mathtt{X}, W_1, W \rangle$
$\langle NP, \mathtt{john}, [John|W], W \rangle \rightarrow John$
$\langle NP, \mathtt{mary}, [Mary|W], W \rangle \rightarrow Mary$
$\langle NP, \mathtt{paris}, [Paris|W], W \rangle \rightarrow Paris$
$\langle V_i, \mathtt{X\hat{}sleep}(\mathtt{X}), [sleeps|W], W \rangle \rightarrow sleeps$
$\langle V_t, \mathtt{X\hat{}Y\hat{}see}(\mathtt{X},\mathtt{Y}), [see|W], W \rangle \rightarrow sees$
$\langle P, \mathtt{X\hat{}in}(\mathtt{X}), [in|W], W \rangle \rightarrow in$
$\langle AdvP, \mathtt{today}, [today|W], W \rangle \rightarrow today$
$\langle QM, \mathtt{ynq}, [?|W], W \rangle \rightarrow ?$

Figure 3: Sample grammar with semantics

A simple semantic-head-driven generator might work as follows: Given a grammar symbol and a piece of logical form, the generator looks for a non-chain rule with the given semantics. The constituents of the RHS of that rule are then generated recursively, after which the LHS is connected to the given grammar symbol using chain rules. At each application of a chain rule, the rest of the RHS constituents, i.e., the non-head constituents, are generated recursively. The particular combination of connecting chain rules used is often referred to as a chain. The generator starts off with the top symbol of the grammar and the logical form corresponding to the string that is to be generated.

The inherent problem with the SHDG algorithm is that each rule combination is tried in turn, while the possibilities of prefiltering are rather limited, leading to a large amount of spurious search. The generation algorithm presented in the current article does not suffer from this problem; what the new algorithm in effect does is to process all chains from a particular set of grammar symbols down to some particular piece of logical form in parallel before any rule is applied, rather than to construct and try each one separately in turn.

## 4  Grammar Inversion

Before we can invert the grammar, we must put it in normal form. We will use a variant of chain and non-chain rules, namely functor-introducing rules corresponding to non-chain rules, and argument-filling rules corresponding to chain rules. The inversion step is based on the assumption that there are no other types of rules.

Since the generator will work by recursive descent through the logical form, we wish to rearrange the grammar so that arguments are generated together with their functors. To this end we introduce another difference list $A_0$ and $A$ to pass down the arguments introduced

**Functor-introducing rules**
$\langle S, \mathtt{mod}(\mathtt{X},\mathtt{Y}), W_0, W, \epsilon, \epsilon \rangle \rightarrow$  1
  $\langle S, \mathtt{X}, W_0, W_1, \epsilon, \epsilon \rangle \ \langle QM, \mathtt{Y}, W_1, W, \epsilon, \epsilon \rangle$
$\langle VP, \mathtt{X\hat{}mod}(\mathtt{Y},\mathtt{Z}), W_0, W, A_0, A \rangle \rightarrow$  3
  $\langle VP, \mathtt{X\hat{}Y}, W_0, W_1, A_0, A \rangle \ \langle AdvP, \mathtt{Z}, W_1, W, \epsilon, \epsilon \rangle$
$\langle VP, \mathtt{X\hat{}mod}(\mathtt{Y},\mathtt{Z}), W_0, W, A_0, A \rangle \rightarrow$  4
  $\langle VP, \mathtt{X\hat{}Y}, W_0, W_1, A_0, A \rangle \ \langle PP, \mathtt{Z}, W_1, W, \epsilon, \epsilon \rangle$
$\langle NP, \mathtt{john}, [John|W], W, A, \epsilon \rangle \rightarrow A$
$\langle NP, \mathtt{mary}, [Mary|W], W, A, \epsilon \rangle \rightarrow A$
$\langle NP, \mathtt{paris}, [Paris|W], W, A, \epsilon \rangle \rightarrow A$
$\langle V_i, \mathtt{X\hat{}sleep}(\mathtt{X}), [sleeps|W], W, A, \epsilon \rangle \rightarrow A$
$\langle V_t, \mathtt{X\hat{}Y\hat{}see}(\mathtt{X},\mathtt{Y}), [see|W], W, A, \epsilon \rangle \rightarrow A$
$\langle P, \mathtt{X\hat{}in}(\mathtt{X}), [in|W], W, A, \epsilon \rangle \rightarrow A$
$\langle AdvP, \mathtt{today}, [today|W], W, A, \epsilon \rangle \rightarrow A$
$\langle QM, \mathtt{ynq}, [?|W], W, A, \epsilon \rangle \rightarrow A$

**Argument-filling rules**
$\langle S, \mathtt{Y}, W_0, W, \epsilon, \epsilon \rangle \rightarrow$  2
  $\langle VP, \mathtt{X\hat{}Y}, W_1, W, [\langle NP, \mathtt{X}, W_0, W_1 \rangle], \epsilon \rangle$
$\langle VP, \mathtt{X}, W_0, W, A_0, A \rangle \rightarrow$  5
  $\langle V_i, \mathtt{X}, W_0, W, A_0, A \rangle$
$\langle VP, \mathtt{Y}, W_0, W, A_0, A \rangle \rightarrow$  6
  $\langle V_t, \mathtt{X\hat{}Y}, W_0, W_1, [\langle NP, \mathtt{X}, W_1, W \rangle | A_0], A \rangle$
$\langle PP, \mathtt{Y}, W_0, W, A_0, A \rangle \rightarrow$  7
  $\langle P, \mathtt{X\hat{}Y}, W_0, W_1, [\langle NP, \mathtt{X}, W_1, W \rangle | A_0], A \rangle$

Figure 4: Sample grammar in normal form

by argument-filling rules to the corresponding functor-introducing rules. Here the latter rules are assumed to be lexical, following the tradition in GPSG where the presence of the SUBCAT feature implies a preterminal grammar symbol (see [Gazdar et al. 1985], p. 33), but this is really immaterial for the algorithm.

The grammar of Figure 3 is shown in normal form in Figure 4. The grammar is compiled into this form by inspecting the flow of arguments through the logical forms of the constituents of each rule. In the functor-introducing rules, the RHS is rearranged to mirror the argument order of the LHS logical form. The argument-filling rules have only one RHS constituent — the semantic head — and the rest of the original RHS constituents are added to the argument list of the head constituent. Note, for example, how the NP is added to the argument list of the VP in Rule 2, or to the argument list of the P in Rule 7. This is done automatically, although currently, the exact flow of arguments is specified manually.

We assume that there are no purely argument-filling cycles. For rules that actually fill in arguments, this is obviously impossible, since the number of arguments decreases strictly. For the slightly degenerate case of argument-filling rules which only pass along the logical form, such as the $\langle VP, \mathtt{X} \rangle \rightarrow \langle V_i, \mathtt{X} \rangle$ rule, this is equivalent to the off-line parsability requirement, see [Kaplan & Bresnan 1982], pp. 264–266.[2] We require this in order to avoid an infinite number of chains, since each possible chain will be expanded out in the inversion step. Since subcategorization lists of verbs are bounded in length, PATR II style VP rules do not pose a serious problem,

---
[2]If the RHS $V_i$ were a VP, we would have a purely argument-filling cycle of length 1.

$$\langle S, \mathtt{mod(X,Y)}, W_0, W, \epsilon, \epsilon \rangle \rightarrow$$
$$\langle S, \mathtt{X}, W_0, W_1, \epsilon, \epsilon \rangle \ \langle QM, \mathtt{Y}, W_1, W, \epsilon, \epsilon \rangle$$
$$\langle S, \mathtt{X\hat{}mod(Y,Z)}, W_0, W, \epsilon, \epsilon \rangle \rightarrow$$
$$\langle VP, \mathtt{X\hat{}Y}, W_1, W_2, [\langle NP, \mathtt{X}, W_0, W_1 \rangle], \epsilon \rangle$$
$$\langle AdvP, \mathtt{Z}, W_2, W, \epsilon, \epsilon \rangle$$
$$\langle S, \mathtt{X\hat{}mod(Y,Z)}, W_0, W, \epsilon, \epsilon \rangle \rightarrow$$
$$\langle VP, \mathtt{X\hat{}Y}, W_1, W_2, [\langle NP, \mathtt{X}, W_0, W_1 \rangle], \epsilon \rangle$$
$$\langle PP, \mathtt{Z}, W_2, W, \epsilon, \epsilon \rangle$$
$$\langle VP, \mathtt{X\hat{}mod(Y,Z)}, W_1, W, [\langle NP, \mathtt{X}, W_0, W_1 \rangle], \epsilon \rangle \rightarrow$$
$$\langle VP, \mathtt{X\hat{}Y}, W_1, W_2, [\langle NP, \mathtt{X}, W_0, W_1 \rangle], \epsilon \rangle$$
$$\langle AdvP, \mathtt{Z}, W_2, W, \epsilon, \epsilon \rangle$$
$$\langle VP, \mathtt{X\hat{}mod(Y,Z)}, W_1, W, [\langle NP, \mathtt{X}, W_0, W_1 \rangle], \epsilon \rangle \rightarrow$$
$$\langle VP, \mathtt{X\hat{}Y}, W_1, W_2, [\langle NP, \mathtt{X}, W_0, W_1 \rangle], \epsilon \rangle$$
$$\langle PP, \mathtt{Z}, W_2, W, \epsilon, \epsilon \rangle$$
$$\langle S, \mathtt{sleep(X)}, W_0, W, \epsilon, \epsilon \rangle \rightarrow \langle NP, \mathtt{X}, W_0, [sleeps|W], \epsilon, \epsilon \rangle$$
$$\langle VP, \mathtt{X\hat{}sleep(X)}, [sleeps|W], W, [\langle NP, \mathtt{X}, W_0, [sleeps|W] \rangle], \epsilon \rangle$$
$$\rightarrow \langle NP, \mathtt{X}, W_0, [sleeps|W], \epsilon, \epsilon \rangle$$
$$\langle S, \mathtt{see(X,Y)}, W_0, W, \epsilon, \epsilon \rangle \rightarrow$$
$$\langle NP, \mathtt{X}, W_1, W, \epsilon, \epsilon \rangle \ \langle NP, \mathtt{Y}, W_0, [sees|W_1], \epsilon, \epsilon \rangle$$
$$\langle VP, \mathtt{Y\hat{}see(X,Y)}, [sees|W_0], W, [\langle NP, \mathtt{Y}, W_1, [sees|W_0] \rangle], \epsilon \rangle$$
$$\rightarrow \langle NP, \mathtt{X}, W_0, W, \epsilon, \epsilon \rangle \ \langle NP, \mathtt{Y}, W_1, [sees|W_0], \epsilon, \epsilon \rangle$$
$$\langle PP, \mathtt{X\hat{}in(X)}, [in|W_0], W, \epsilon, \epsilon \rangle \rightarrow \langle NP, \mathtt{X}, W_0, W, \epsilon, \epsilon \rangle$$
$$\langle NP, \mathtt{john}, [John|W], W, \epsilon, \epsilon \rangle \rightarrow \epsilon$$
$$\langle NP, \mathtt{mary}, [Mary|W], W, \epsilon, \epsilon \rangle \rightarrow \epsilon$$
$$\langle NP, \mathtt{paris}, [Paris|W], W, \epsilon, \epsilon \rangle \rightarrow \epsilon$$
$$\langle AdvP, \mathtt{today}, [today|W], W, \epsilon, \epsilon \rangle \rightarrow \epsilon$$
$$\langle QM, \mathtt{ynq}, [?|W], W, \epsilon, \epsilon \rangle \rightarrow \epsilon$$

Figure 5: Inverted sample grammar

which on the other hand the "adjunct-as-argument" approach taken in [Bouma & van Noord 1994] may do. However, this problem is common to a number of other generation algorithms, including the SHDG algorithm.

Let us return to the scenario for the SHDG algorithm given at the end of Section 3: We have a piece of logical form and a grammar symbol, and we wish to connect a non-chain rule with this particular logical form to the given grammar symbol through a chain. We will generalize this scenario just slightly to the case where a set of grammar symbols is given, rather than a single one.

Each inverted rule will correspond to a particular chain of argument-filling (chain) rules connecting a functor-introducing (non-chain) rule introducing this logical form to a grammar symbol in the given set. The arguments introduced by this chain will be collected and passed down to the functors that consume them in order to ensure that each of the inverted rules has a RHS matching the structure of the LHS logical form. The normalized sample grammar of Figure 4 will result in the inverted grammar of Figure 5. Note how the right-hand sides reflect the argument structure of the left-hand-side logical forms. As mentioned previously, the collected arguments are currently assumed to correspond to functors introduced by lexical entries, but the procedure can readily be modified to accommodate grammar rules with a non-empty RHS, where some of the arguments are consumed by the LHS logical form.

The grammar inversion step is combined with the LR-compilation step. This is convenient for several reasons: Firstly, the termination criteria and the database maintenance issues are the same in both steps. Secondly,

State 1
$$\langle S', \mathtt{f(X)}, \epsilon, \epsilon \rangle \ \Rightarrow \ \cdot \langle S, \mathtt{X}, \epsilon, \epsilon \rangle$$

State 2
$$\langle S, \mathtt{mod(X,Y)}, \epsilon, \epsilon \rangle \ \Rightarrow \ \cdot \langle S, \mathtt{X}, \epsilon, \epsilon \rangle \ \langle QM, \mathtt{Y}, \epsilon, \epsilon \rangle$$
$$\langle S, \mathtt{mod(Y,Z)}, \epsilon, \epsilon \rangle \ \Rightarrow \ \cdot \langle VP, \mathtt{X\hat{}Y}, [\langle NP, \mathtt{X} \rangle], \epsilon \rangle \ \langle AdvP, \mathtt{Z}, \epsilon, \epsilon \rangle$$
$$\langle S, \mathtt{mod(Y,Z)}, \epsilon, \epsilon \rangle \ \Rightarrow \ \cdot \langle VP, \mathtt{X\hat{}Y}, [\langle NP, \mathtt{X} \rangle], \epsilon \rangle \ \langle PP, \mathtt{Z}, \epsilon, \epsilon \rangle$$

State 3
$$\langle S, \mathtt{mod(X,Y)}, \epsilon, \epsilon \rangle \ \Rightarrow \ \cdot \langle S, \mathtt{X}, \epsilon, \epsilon \rangle \ \langle QM, \mathtt{Y}, \epsilon, \epsilon \rangle$$
$$\langle S, \mathtt{mod(Y,Z)}, \epsilon, \epsilon \rangle \ \Rightarrow \ \cdot \langle VP, \mathtt{X\hat{}Y}, [\langle NP, \mathtt{X} \rangle], \epsilon \rangle \ \langle AdvP, \mathtt{Z}, \epsilon, \epsilon \rangle$$
$$\langle S, \mathtt{mod(Y,Z)}, \epsilon, \epsilon \rangle \ \Rightarrow \ \cdot \langle VP, \mathtt{X\hat{}Y}, [\langle NP, \mathtt{X} \rangle], \epsilon \rangle \ \langle PP, \mathtt{Z}, \epsilon, \epsilon \rangle$$
$$\langle VP, \mathtt{X\hat{}mod(Y,Z)}, [\langle NP, \mathtt{X} \rangle], \epsilon \rangle \ \Rightarrow$$
$$\cdot \langle VP, \mathtt{X\hat{}Y}, [\langle NP, \mathtt{X} \rangle], \epsilon \rangle \ \langle AdvP, \mathtt{Z}, \epsilon, \epsilon \rangle$$
$$\langle VP, \mathtt{X\hat{}mod(Y,Z)}, [\langle NP, \mathtt{X} \rangle], \epsilon \rangle \ \Rightarrow$$
$$\cdot \langle VP, \mathtt{X\hat{}Y}, [\langle NP, \mathtt{X} \rangle], \epsilon \rangle \ \langle PP, \mathtt{Z}, \epsilon, \epsilon \rangle$$

Figure 6: The first three generation states

since the LR-compilation step employs a top-down rule-invocation scheme, this will ensure that the arguments are passed down to the corresponding functors. In fact, invoking inverted grammar rules merely requires first invoking a chain of argument-filling rules and then terminating it with a functor-introducing rule.

## 5 LR Compilation for Generation

Just as when compiling LR-parsing tables, the compiler operates on sets of dotted items. Each item consists of a partially processed inverted grammar rule, with a dot marking the current position. Here the current position is an argument position of the LHS logical form, rather than some position in the input string.

New states are induced from old ones: For the indicated argument position, a possible logical form is selected and the dot is advanced one step in all items where this particular logical form can occur in the current argument position, and the resulting new items constitute a new state. All possible grammar symbols that can occur in the old argument position and that can have this logical form are then collected. From these, all rules with a matching LHS are invoked from the inverted grammar. Each such rule will give rise to a new item where the dot marks the first argument position, and the set of these new items will constitute another new state. If a new set of items is constructed that is more specific than an existing one, then this search branch is abandoned and the recursion terminates. If it on the other hand is more general, then it replaces the old one.

The state-construction phase starts off by creating an initial set consisting of a single dummy item with a dummy top grammar symbol and a dummy top logical form, corresponding to a dummy inverted grammar rule. In the sample grammar, this would be the rule $\langle S', \mathtt{f(X)}, W_0, W, \epsilon, \epsilon \rangle \rightarrow \langle S, \mathtt{X}, W_0, W, \epsilon, \epsilon \rangle$. The dot is at the beginning of the rule, selecting the first and only argument. The rest of the states are induced from this one. The first three states resulting from the inverted grammar of Figure 5 are shown in Figure 6, where the difference lists representing the word strings are omitted.

The sets of items are used to compile the generation tables in the same way as is done for LR parsing. The goto entries correspond to transiting from one argument of a term to the next, and thus advancing the dot one step. The reductions correspond to applying the rules of items that have the dot at the end of the RHS, as is the case when LR parsing. There is no obvious analogy to the shift action — the closest thing would be the descend actions transiting from a functor to one of its arguments.

Note that there is no need to include the logical form of each lexicon entry in the generation tables. Instead, a typing of the logical forms can be introduced, and a representative of each type used in the actual tables, rather than the individual logical forms. This decreases the size of the tables drastically. For example, there is no point in distinguishing the states reached by traversing john, mary and paris, apart from ensuring that the correct word is added to the output word-string. This is accomplished much in the same way as preterminals, rather than individual words, figure in LR-parsing tables.

## 6  A New Generation Algorithm

The generator works by recursive descent through the logical form while transiting between internal states. It is driven by the descend, goto and reduce tables. A pushdown stack is used to store intermediate constituents.

When generating a word string, the current state and logical form determine a transition to a new state, corresponding to the first argument of the logical form, through the *descend* table. A substring is generated recursively from the argument logical form, and this constituent is pushed onto the stack. The argument logical form, together with the new current state, determine a transition to the next state through the *goto* table. The next state corresponds to the next argument of the original logical form, and another substring is generated from this argument logical form, etc. When no more arguments remain, an inverted grammar rule is selected nondeterministically by the *reduce* table and applied to the top portion of the stack, constructing a word string corresponding to the original logical form and completing this generation cycle.[3]

The logical form can be inspected down to an arbitrary depth of recursion when compiling the sets of items, and this parameter can be varied. This is closely related to the use of lookahead symbols in an LR parser; increasing the depth is analogous to increasing the number of lookahead symbols. The amount of semantic lookahead is reflected in the goto and descend entries. The key parameter influencing the generation speed is the amount of nondeterminism in each "reductive state", i.e., each state where the dot is at the end of some rule. Increased semantic lookahead will split potential nondeterminism in the resulting reductive states into distinct sets of items, yielding reductive states with less nondeterminism.

No semantic lookahead would mean only taking the functor of the logical form into consideration, and in the

---

[3]This is a bottom-up rule invocation scheme. It could easily be modified so that a rule is instead applied *before* constructing the substrings recursively, resulting in a top-down rule-invocation scheme, which might be a good idea in conjunction with semantic lookahead, see the following.

```
descend(1, mod(mod(_,_),ynq), 2A).
descend(1, mod(see(_,_),ynq), 2B).
descend(1, mod(sleep(_),ynq), 2C).
```

State 2A
$\langle S, \text{mod}(\text{mod}(X,Y), \text{ynq}), \epsilon, \epsilon \rangle \Rightarrow$
  $\cdot \langle S, \text{mod}(X,Y), \epsilon, \epsilon \rangle \langle QM, \text{ynq}, \epsilon, \epsilon \rangle$

State 2B
$\langle S, \text{mod}(\text{see}(X,Y), \text{ynq}), \epsilon, \epsilon \rangle \Rightarrow$
  $\cdot \langle S, \text{see}(X,Y), \epsilon, \epsilon \rangle \langle QM, \text{ynq}, \epsilon, \epsilon \rangle$

State 2C
$\langle S, \text{mod}(\text{sleep}(X), \text{ynq}), \epsilon, \epsilon \rangle \Rightarrow$
  $\cdot \langle S, \text{sleep}(X), \epsilon, \epsilon \rangle \langle QM, \text{ynq}, \epsilon, \epsilon \rangle$

Figure 7: Alternative generation states

```
descend(1, mod(_,ynq), 2).
```

State 2
$\langle S, \text{mod}(X, \text{ynq}), \epsilon, \epsilon \rangle \Rightarrow \cdot \langle S, X, \epsilon, \epsilon \rangle \langle QM, \text{ynq}, \epsilon, \epsilon \rangle$

Figure 8: Alternative alternative generation states

running example, a typical action table entry would be descend(1,mod(_,_),2).[4] This would mean that the generator would operate on State 2 of Figure 6 when generating from the first argument of the mod(_,_) term, and both the $S$ alternative and the (merged) $VP$ alternative(s) would be attempted nondeterministically.

By taking the arguments of the logical form into account, the degree of nondeterminism can be reduced, and for the sample grammar used throughout this article, it is eliminated completely. In the example, if the second argument of the mod(_,_) term is ynq, then only the $S$ alternative will be considered when generating from the first argument, since the relevant states and descend entries will be those of Figure 7.

The optimal depth may vary for each individual table entry, and even within it, and a scheme has been devised to automatically find such an optimum by inspecting the number of items left in each reductive state. The scheme employs a greedy algorithm with iterative deepening to this end. In the running example, the first argument of mod(_,_) contributes no important information when descending from State 1, while the second one does. The scheme correctly finds the optimal depths when transiting from State 1, resulting in the State 2 and descend entry of Figure 8. This is described in detail elsewhere.

## 7  Summary and Discussion

The proposed algorithm is an improvement on the semantic-head-driven generation algorithm that allows "functor merging", i.e., enables processing various grammar rules, or rule combinations, that introduce the same semantic structure simultaneously, thereby greatly reducing the search space. The algorithm proceeds by recursive descent through the logical form, and using the terminology of the SHDG algorithm, what the new algorithm in effect does is to process all chains from a particular set of grammar symbols down to some particular piece of logical form in parallel until a reduction is attempted, rather than to construct and try each one separately in turn. This requires a grammar-inversion technique that is fundamentally different from techniques such as the essential-argument algorithm, see the following, since it must display the grammar from the point of view of the logical form, rather than from that of the word string. LR-compilation techniques accomplish the functor merging by compiling the inverted grammar into a set of generation tables.

---

[4]Here "_" denotes a don't-care variable.

The set of applicable reductions can be reduced by using more semantic lookahead, at the price of a larger number of internal states, and there is in general a tradeoff between the size of the resulting generation tables and the amount of nondeterminism when reducing. The employed amount of semantic lookahead can be varied, and a scheme has been devised and tested that automatically determines appropriate tradeoff points, optionally based on a collection of training examples.

The grammar inversion rearranges the grammar as a whole according to the functor-argument structure of the logical forms. Other inversion schemes, such as the essential-argument algorithm, see [Strzalkowski 1990] or the direct-inversion approach, see [Minnen et al. Forthcoming], are mainly concerned with locally rearranging the order of the RHS constituents of individual grammar rules by examining the flow of information through these constituents, to ensure termination and increase efficiency. Although this can occasionally change the set of RHS symbols in a rule, it is done to these ends, rather than to reflect the functor-argument structure.

Some hand editing is necessary when preparing the grammar for the inversion step, but it is limited to specifying the flow of arguments in the grammar rules. Furthermore, this could potentially be fully automated.

Although the sample grammar used throughout the article is essentially context-free, there is nothing in principle that restricts the method to such grammars. In fact, the method could be extended to grammars employing complex feature structures as easily as the LR-parsing scheme itself, see for example [Nakazawa 1991], and this is currently being done.

The method has been implemented and applied to much more complex grammars than the simple one used as an example in this article, and it works excellently. Although these grammars are still too naive to form the basis of a serious empirical evaluation lending substantial experimental support to the method as a whole, it should be obvious from the algorithm itself that the reduction in search space compared to the SHDG algorithm is most substantial. Nonetheless, such an evaluation is a top-priority item on the future-work agenda.

## Acknowledgements

I wish to thank greatly Gregor Erbach, Jussi Karlgren, Manny Rayner, Hans Uszkoreit, Mats Wirén and the anonymous reviewers of ACL, EACL and IJCAI for valuable comments and suggestions to improvements on draft and previous versions of this article and other related publications. Special credit is due to Kristina Striegnitz, who assisted with the implementation of the system.